\documentclass[10pt,twocolumn,letterpaper]{article}

\usepackage[pagenumbers]{cvpr} 

\definecolor{cvprblue}{rgb}{0.21,0.49,0.74}
\usepackage[pagebackref,breaklinks,colorlinks,allcolors=cvprblue]{hyperref}

\usepackage{graphicx}
\usepackage{booktabs}
\usepackage{wrapfig}
\usepackage{bm}
\usepackage{multirow}

\def\paperID{255} 
\def\confName{3DV\xspace}
\def\confYear{2027\xspace}

\title{Elastic Triangle Splatting}

\author{
Tian Shi$^{1,2,*}$ \quad
Shenhan Qian$^{1,2,*}$ \quad
Daniel Cremers$^{1,2}$
\\[8pt]
\begin{tabular}[t]{c c c}%
    $^1$Technical University of Munich &
    $^2$MCML
\end{tabular}
}

\begin{document}
\twocolumn[{
\renewcommand\twocolumn[1][]{#1}
\maketitle
\begin{center}
    \centering
    \captionsetup{type=figure}
    \includegraphics[width=1\textwidth]{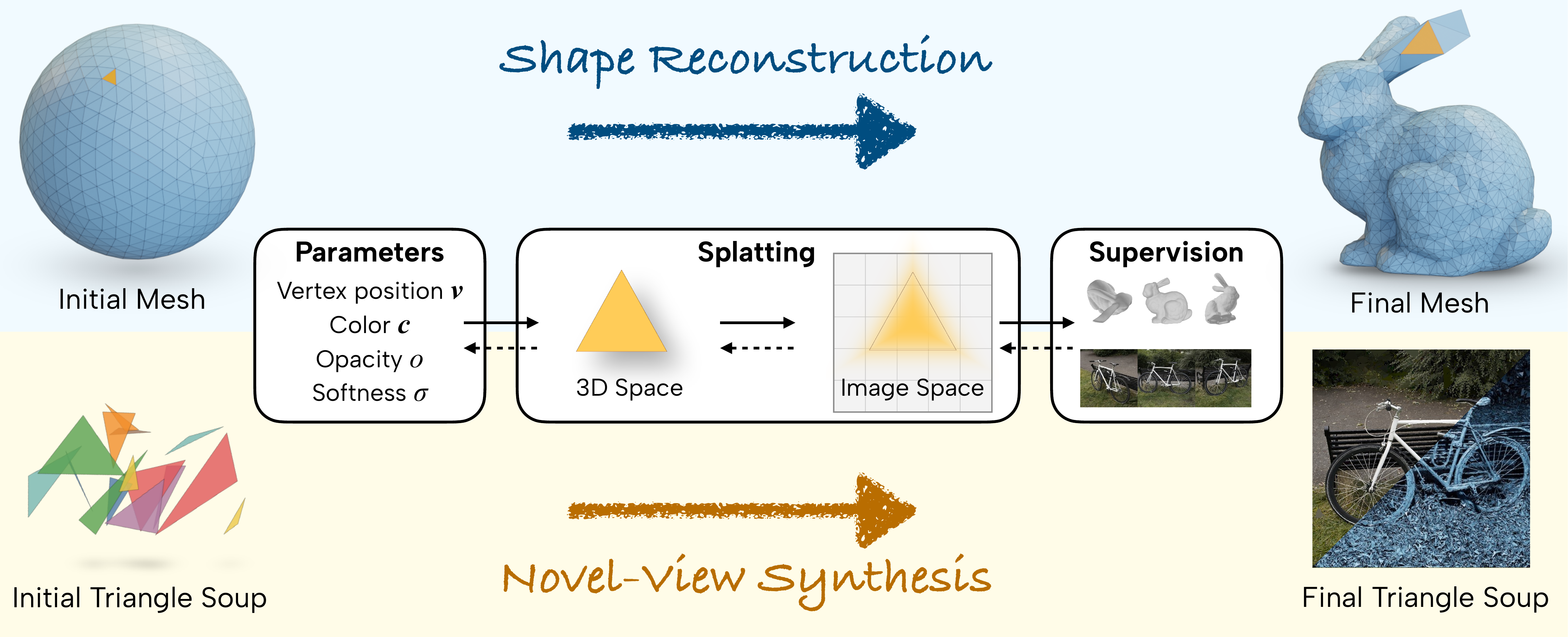}
    \caption{\textbf{Elastic Triangle Splatting} explores kernel properties through two complementary tasks. \textit{Shape reconstruction} (top) deforms a spherical triangle mesh into an accurate surface. \textit{Novel-view synthesis} (bottom) optimizes an unstructured triangle soup for photorealistic free-viewpoint rendering. Triangles are splatted onto the image plane with a soft kernel and optimized using supervision from posed RGB images. Each triangle is parameterized by vertex positions $\bm{v}$, face color $\bm{c}$, face opacity $o$, and edge softness $\sigma$.}
    \label{fig:teaser}
\end{center}
}]

\renewcommand{\thefootnote}{}%
\footnotetext{$^*$Equal contribution.}
\footnotetext{Project page: \url{https://shenhanqian.github.io/ets}}
\renewcommand{\thefootnote}{\arabic{footnote}}

\begin{abstract}
While neural rendering methods such as 3D Gaussian Splatting achieve remarkable visual fidelity, traditional polygonal meshes remain the backbone of established graphics pipelines. Triangle splatting bridges this gap by optimizing triangle primitives as differentiable splats, producing representations that are closer to mesh-based workflows. Central to these methods is the kernel function that softens triangle boundaries to propagate gradients to vertex positions. Existing triangle splatting methods make inconsistent choices of kernel functions, and analysis of these kernels' optimization behavior has been limited to unstructured triangle soups for novel-view synthesis. In this work, we consider triangle splatting as a generic tool for photometric optimization, comparing kernel properties through two complementary tasks: mesh optimization for shape reconstruction and triangle soup optimization for novel-view synthesis. Along with the analysis, we introduce an elastic kernel function that features bilateral gradient support across the boundary and an adaptive boundary value, which are shown to be essential for robust optimization. Under isolated comparison, our elastic kernel outperforms existing kernels on shape reconstruction and in the majority of novel-view synthesis benchmarks, demonstrating the importance of kernel design in the effectiveness and versatility of triangle splatting.
\end{abstract}
    
\section{Introduction}
\label{sec:intro}

Neural rendering has fundamentally reshaped 3D scene reconstruction, with methods such as Neural Radiance Fields (NeRF)~\cite{mildenhall2020nerf} and 3D Gaussian Splatting (3DGS)~\cite{kerbl20233d} setting new benchmarks for novel-view synthesis (NVS). Yet these volumetric and point-based representations operate largely outside traditional graphics pipelines: Gaussian primitives render effectively but do not tile a surface without overlap or gaps, making deformation, simulation, and mesh editing difficult, while standard pipelines revolve almost exclusively around the polygonal mesh. Consequently, integrating neural scene representations into existing AR/VR frameworks, physics engines, and game engines often requires lossy and computationally expensive conversion steps~\cite{chen2022mobilenerf,yu2024gof}.

\begin{figure}[t]
  \centering
  \includegraphics[width=0.5\linewidth]{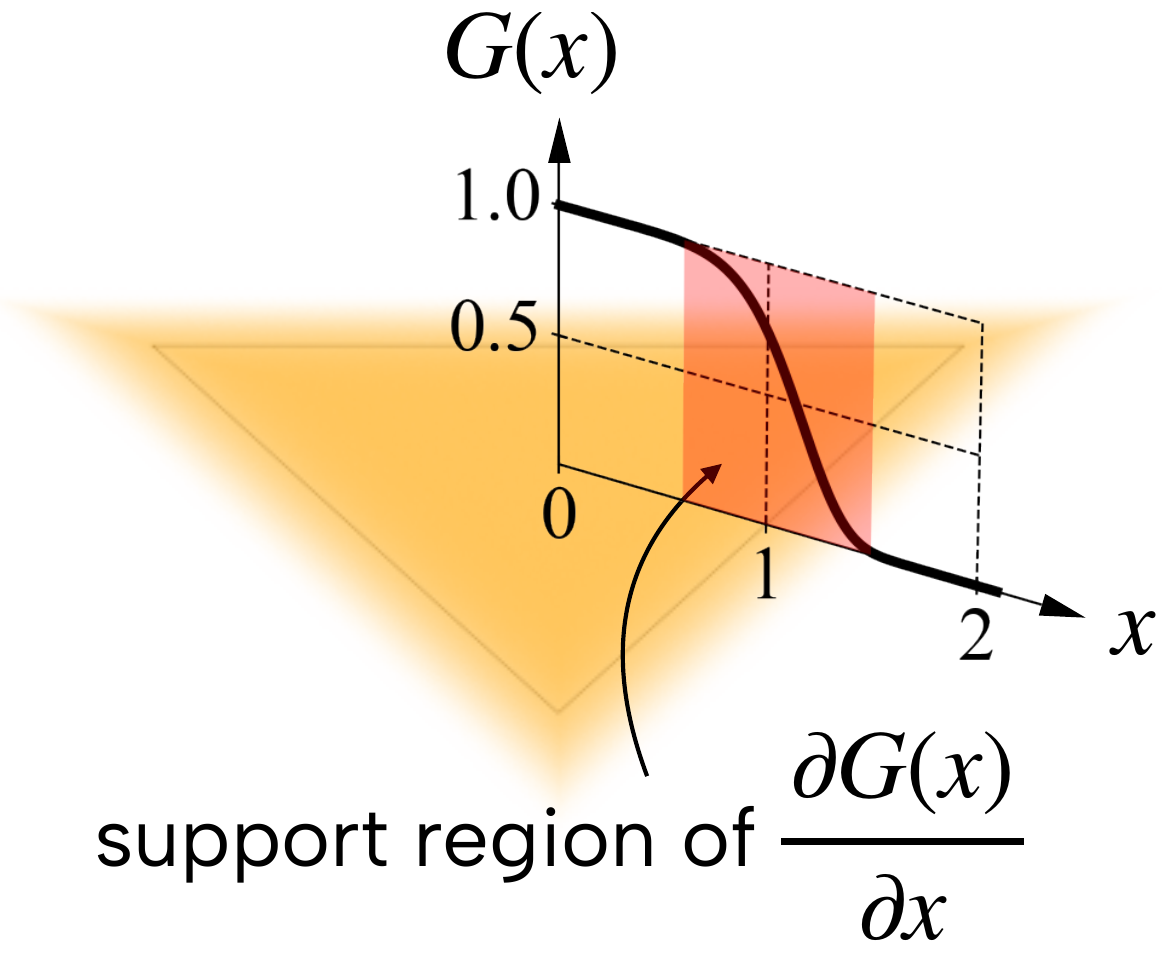}
  \caption{A triangle with a softened boundary.}
  \label{fig:softtri}
\end{figure}

This disconnect has sparked a recent ``triangle comeback''~\cite{held2025triangle, sheng20252d}. Triangle splatting methods seek to bridge this gap by adopting the triangle---the fundamental primitive of graphics---as the core unit of differentiable optimization. By treating triangles as differentiable splats, these methods aim to combine the rendering quality of 3DGS with the structural compatibility of traditional meshes. Although custom splatting kernels are still required during optimization, the final optimized triangles can be rendered by standard rasterization without converting from an intermediate volumetric primitive. Central to this approach is the kernel function, a mathematical formulation designed to ``soften'' the sharp boundaries of triangles as shown in \cref{fig:softtri}. This softening is critical: it enables gradients from image-space losses to flow back to vertex positions, making the otherwise discrete geometry of a mesh amenable to gradient-based optimization.

Despite the promise of triangle-based splatting, existing methods make different kernel choices that lead to different optimization behavior. Triangle Splatting~\cite{held2025triangle} softens a triangle only on its inner side, while 2DTS~\cite{sheng20252d} adopts a bilateral kernel that spreads color to both sides of the triangle boundary (see \cref{fig:func} for visual illustrations). These designs already demonstrate that the kernel is not a minor implementation detail. However, their behavior is usually evaluated within different pipelines and tasks, making it difficult to isolate which kernel properties matter most for geometry optimization. It is then natural to ask: how much does the choice of kernel function affect optimization dynamics and convergence? This motivates a focused comparison of kernel properties such as support, smoothness, decay, and boundary value---characteristics that govern the trade-off between wide-reaching gradients early in training and sharp, artifact-free boundaries at convergence.

Evaluating kernel functions in NVS alone is insufficient, however, because the favorable point-cloud initialization from structure-from-motion and the adaptive densification strategy can mask the kernel's intrinsic behavior. To isolate the kernel's contribution, we also study mesh optimization: a controlled setting in which a fixed-topology spherical mesh is the sole initialization. This enables a direct comparison against classic differentiable rasterization baselines such as Nvdiffrast~\cite{laine2020modular}, cleanly attributing performance differences to the kernel design itself.

In this paper, we revisit kernel functions for triangle-based splatting under a common formulation, examining their mathematical properties---support, smoothness, decay, and boundary value---through a unified framework (see \cref{tab:kernel_axes} for a summary of existing designs). Our study spans two complementary tasks: (1) \textit{Mesh Optimization for Shape Reconstruction}, where a connected triangle mesh deforms to recover accurate geometric surfaces, and (2) \textit{Triangle Soup Optimization for Novel-View Synthesis}, where thousands of independent triangles are jointly optimized for photometric fidelity.

\begin{table}[t]
\caption{\textbf{Kernel design axes.} Existing triangle kernels cover either interior-only support or bilateral support with fixed boundary weight. Our kernel keeps bilateral gradients while making the boundary value adaptive through $\sigma$.}
\label{tab:kernel_axes}
\centering
\footnotesize
\setlength{\tabcolsep}{2pt}
\begin{tabular}{@{}lcccc@{}}
\toprule
Kernel & Support & $G(1)$ & Exterior grad. & Main limitation \\
\midrule
Triangle Splatting~\cite{held2025triangle} & inside & 0 & no & dead zones \\
2DTS~\cite{sheng20252d} & bilateral & fixed & yes & boundary bias \\
Ours & bilateral & adaptive & yes & -- \\
\bottomrule
\end{tabular}
\end{table}

Our analysis identifies an \emph{elastic} behavior as a useful kernel property: the kernel provides wide-reaching gradients on both sides of the triangle boundary to facilitate large-scale geometric movement, while allowing the opacity on the boundary to approach one as optimization converges without producing persistent edge artifacts. Based on this observation, we present \textit{Elastic Triangle Splatting}, a unified differentiable rendering framework for connected meshes and unstructured triangle soups. Across both settings, the elastic kernel improves over the single-sided kernel of Triangle Splatting and the bilateral kernel of 2DTS, leading to state-of-the-art results in shape reconstruction and novel-view synthesis.

Our contributions are summarized as follows:
\begin{itemize}
\item We provide a unified comparison of triangle splatting kernels across connected mesh optimization and unstructured triangle-soup optimization, isolating how support, decay, and boundary value affect gradient flow and convergence.
\item We introduce an elastic kernel with wide bilateral gradient support and an adaptive boundary value, and show that this boundary-adaptive behavior improves both shape reconstruction and novel-view synthesis over existing triangle kernels.
\end{itemize}

\section{Related Work}
\label{sec:related}

\subsection{Differentiable Mesh Rasterization}
Differentiable rendering addresses the challenge of propagating gradients through the inherently discrete rasterization process. Early work includes OpenDR~\cite{loper2014opendr}, which introduced approximate differentiation of the rendering pipeline to enable gradient-based optimization within analysis-by-synthesis frameworks. Neural Mesh Renderer~\cite{kato2018neural} pioneered backpropagation through rasterization by approximating silhouette gradients via a pixel-wise blurring scheme, enabling mesh optimization from 2D image supervision. Subsequent methods, such as SoftRas~\cite{liu2019soft} and PyTorch3D~\cite{ravi2020accelerating}, adopted probabilistic blending to compute analytical gradients more effectively. Nvdiffrast~\cite{laine2020modular} further advanced this direction by enabling high-performance differentiability in the $XY$-plane through analytic anti-aliasing; building on it, nvdiffrec~\cite{munkberg2022nvdiffrec} demonstrates joint reconstruction of triangular geometry, materials, and lighting from multi-view images. Preconditioned optimization~\cite{Nicolet2021Large} enhances the robustness of mesh-based differentiable rendering by mitigating degeneracies without relying on classical Laplacian regularization~\cite{nealen2006laplacian}. DRTK~\cite{pidhorskyi2024rasterized} introduces micro-edges to compute gradients at visibility discontinuities without modifying the forward rasterization pass. Beyond direct mesh optimization, DMTet~\cite{shen2021deep} represents geometry as a deformable tetrahedral grid and resolves topology changes via differentiable isosurface extraction, enabling gradient-based synthesis of high-resolution surfaces.

\subsection{Novel-View Synthesis with Radiance Fields}
NeRF \cite{mildenhall2020nerf} revolutionized 3D reconstruction via implicit MLP-based volumetric encoding. However, extracting high-quality surfaces from NeRF remains challenging due to the inherent ambiguity of volumetric density. To bridge this gap, NeuS \cite{wang2021neus} and VolSDF \cite{yariv2021volume} integrated volumetric rendering with Signed Distance Functions, ensuring surface smoothness and topological consistency through unbiased density transformations. Mip-NeRF 360 \cite{barron2022mipnerf360} extends NeRF to unbounded scenes with anti-aliased scene parameterization. Although works such as Plenoxels \cite{yu_and_fridovichkeil2021plenoxels}, Instant-NGP \cite{muller2022instant}, and TensoRF \cite{chen2022tensorf} significantly accelerate training by replacing MLPs with explicit spatial data structures, these representations still suffer from high per-ray sampling costs, limiting their applicability in real-time graphics.

Departing from implicit fields, 3DGS \cite{kerbl20233d} explicitly anchors differentiable anisotropic Gaussian primitives to point clouds and leverages efficient tile-based CUDA rasterization to achieve ultra-fast, real-time rendering. 3DGS MCMC \cite{kheradmand20243d} reformulates the densification of Gaussian primitives as a Markov Chain Monte Carlo process, providing a principled alternative to the original heuristic cloning and splitting strategy. By constraining volume primitives to 2D planar Gaussians, 2DGS \cite{Huang2DGS2024} achieves more precise geometric reconstruction, with postprocessing to extract meshes from the splats. SuGaR \cite{guedon2024sugar} aligns Gaussian primitives with an underlying mesh surface through regularization, enabling direct and efficient mesh extraction from trained Gaussian scenes. Triangle Splatting \cite{held2025triangle} and 2DTS \cite{sheng20252d} take this further by directly rendering and optimizing triangle soups, making them natively compatible with traditional graphics pipelines. MILo \cite{guedon2025milo} jointly renders Gaussian splats and an extracted mesh simultaneously, exploiting the complementary advantages of both representations. MeshSplatting \cite{Held2025MeshSplatting} builds a two-stage pipeline on top of Triangle Splatting \cite{held2025triangle}: the first stage optimizes a triangle soup for fast convergence, and the second stage refines a connected triangle mesh for accurate geometry.

\section{Method}
\label{sec:method}

\begin{figure*}[t]
  \centering
  \includegraphics[width=0.9\textwidth]{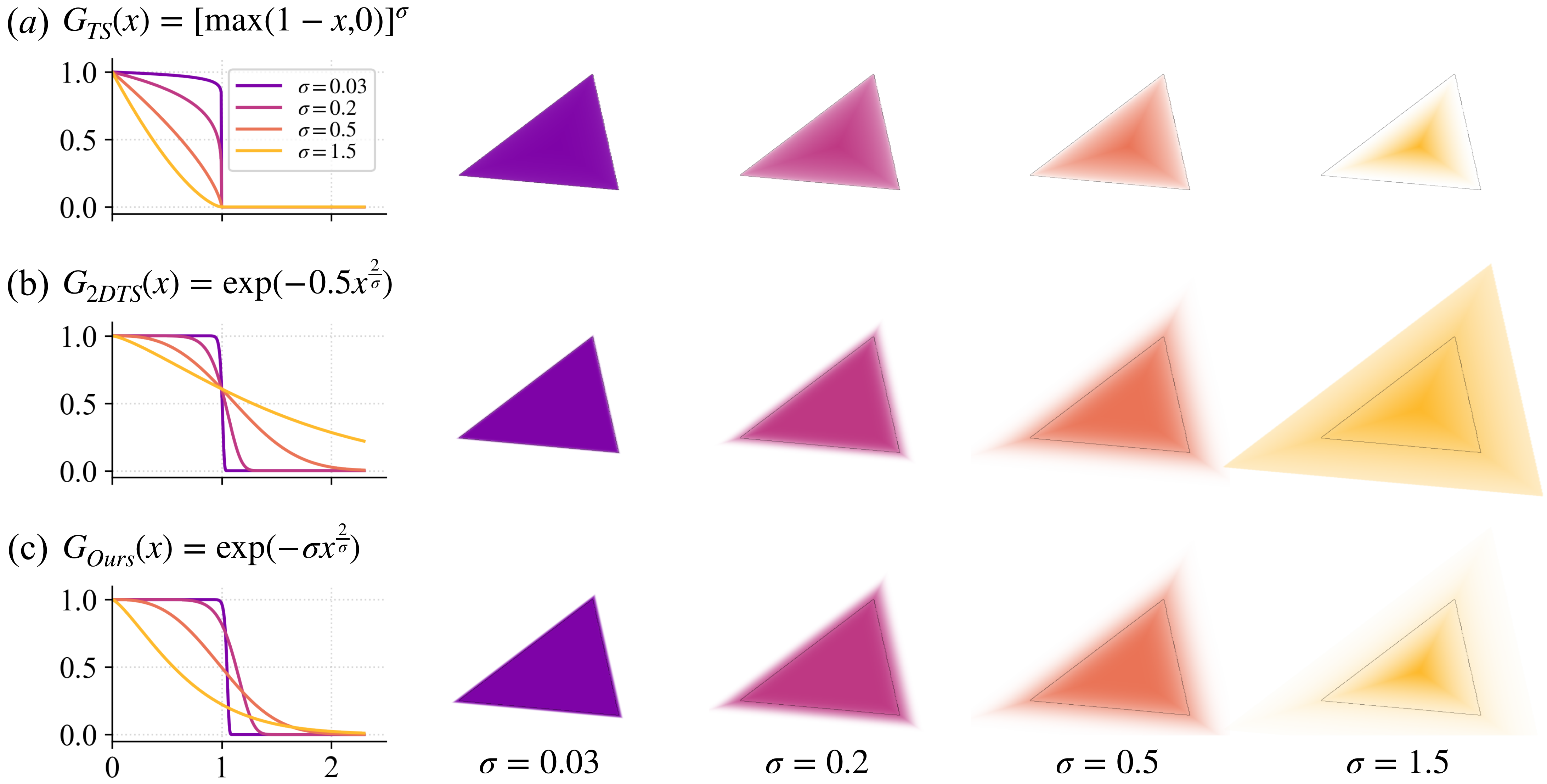}
  \caption{\textbf{Comparison of kernel functions.} (a) Single-sided kernel~\cite{held2025triangle}: non-zero only inside the triangle ($x \leq 1$), providing no exterior gradient and shrinking when softened. (b) Bilateral kernel~\cite{sheng20252d}: extends beyond the triangle boundary, but the boundary weight at $x=1$ is fixed, creating edge artifacts when sharpened and visual expansion when softened. (c) Our elastic kernel: the boundary weight $G(1) = \exp(-\sigma)$ is flexible with respect to $\sigma$, enabling wide gradient support while maintaining a stable visual size.}
  \label{fig:func}
\end{figure*}

We present a unified differentiable rendering framework applicable to two tasks: triangle soup optimization for novel-view synthesis (NVS) and triangle mesh optimization for shape reconstruction. Both tasks share the same core rendering primitive---triangle---but differ in initialization, topology, and regularization. The central design choice governing both is the kernel function, which determines how gradient flow is distributed spatially around each triangle boundary.

\subsection{Triangles as Splatting Primitives}
\label{sec:triangle_splatting}

The shape and appearance of a scene can be represented as a collection of triangles, each equipped with an opacity $o$, three vertex positions $\bm{v}_1,\bm{v}_2,\bm{v}_3 \in \mathbb{R}^3$, per-triangle appearance attributes (colors or spherical harmonic coefficients) $\bm{c}$, and a softness coefficient $\sigma$. Rendering proceeds via a tile-based alpha compositing pipeline adapted from 3DGS~\cite{kerbl20233d}. For each pixel $\bm{p}$, the overlapping triangles are sorted by depth and composited front-to-back:
\begin{equation}
\bm{C}(\bm{p}) = \sum_{k=1}^{K} \bm{c}_k \,\alpha_k(\bm{p}) \prod_{j=1}^{k-1}\left(1 - \alpha_j(\bm{p})\right),
\label{eq:render}
\end{equation}
with $\alpha_k(\bm{p}) = o_k \cdot G\left(x_k(\bm{p})\right)$, where $x_k(\bm{p})$ is the normalized distance from pixel $\bm{p}$ to the center of triangle $k$ on the 2D image plane. The kernel function $G(x)$ acts as a spatial weight that modulates the triangle opacity $o_k$ by the pixel's proximity to the triangle.
Following 2DTS~\cite{sheng20252d}, we define $ x = 1 - 3 \cdot \min(a_{1}, a_{2}, a_{3})$, where $a_{1}, a_{2}, a_{3}$ are the barycentric coordinates of a pixel relative to the triangle. Under this parameterization, $x = 0$ at the triangle centroid, $x = 1$ at the triangle edge, and $x > 1$ in the exterior region (see \cref{fig:softtri} for an illustration). 
Note that there are alternative choices of distance parameterization, such as the signed distance of a pixel to the triangle boundary~\cite{held2025triangle}. However, we empirically find that the choice of distance parameterization has negligible impact on results.
Therefore, we discuss all kernel functions $G(x)$ based on the definition of $x$ to ease comparison of formulas and isolate the effect of kernel design itself. For efficiency of the tile-based rendering pipeline, triangles are culled out for a pixel when $x > 2$, which is equivalent to clamping $G(x)$ to zero for $x > 2$.

\subsection{Kernel Functions for Triangle Rendering}
\label{sec:kernel_func}

The kernel function $G(x) \in [0,1]$ describes the influence of a triangle on a pixel's color based on the distance between them. For an opaque triangle, $G(x)$ is a step function jumping from 1 to 0 when $x=1$. This discrete change creates sharp edges for the triangle but hinders photometric optimization of vertex positions. Specifically, the classical derivative $\frac{\partial G(x)}{\partial x}$ is undefined when $x=1$ and is zero everywhere else, breaking the chain of gradients from the photometric loss to the vertex position:
\begin{equation}
\frac{\partial \bm{C}}{\partial \bm{v}} = 
\frac{\partial \bm{C}}{\partial \alpha}
\frac{\partial \alpha}{\partial G(x)}
\frac{\partial G(x)}{\partial x}
\frac{\partial x}{\partial \bm{v}}
.
\end{equation}
Therefore, selecting the kernel function $G(x)$ is essentially finding a continuous approximation of the step function that provides wide support for the positional gradient.

Here, we compare three representative kernel designs in \cref{fig:func}. They span two axes that are central to triangle splatting optimization: whether the kernel is single-sided or bilateral around the triangle boundary ($x=1$), and whether the boundary value is fixed or can change with the softness parameter. This comparison includes the single-sided kernel used by Triangle Splatting~\cite{held2025triangle}, the bilateral kernel used by 2DTS~\cite{sheng20252d}, and our elastic kernel. We unify the distance parameterization in shape reconstruction experiments to isolate the effect of kernel functions (see details in the appendix (\cref{sec:kernel_unification})).

\paragraph{Single-sided Kernel.}
Triangle Splatting~\cite{held2025triangle} adopts a single-sided kernel function:
\begin{equation}
G(x) = [\max(1-x, 0)]^{\sigma}.
\end{equation}
It always has $G(0)=1$ and $G(1)=0$, with the softness coefficient $\sigma$ adjusting the curve in between. This kernel assigns zero weight for $x > 1$, which seems desirable because pixels outside the physical boundary of a triangle are never influenced. However, it also means that no gradient signal is provided from exterior pixels to vertex positions. Once a triangle boundary retreats past a target surface region, gradient flow vanishes entirely, causing optimization to stall and the triangle to collapse.

\begin{figure}[t]
  \centering
  \includegraphics[width=0.8\linewidth]{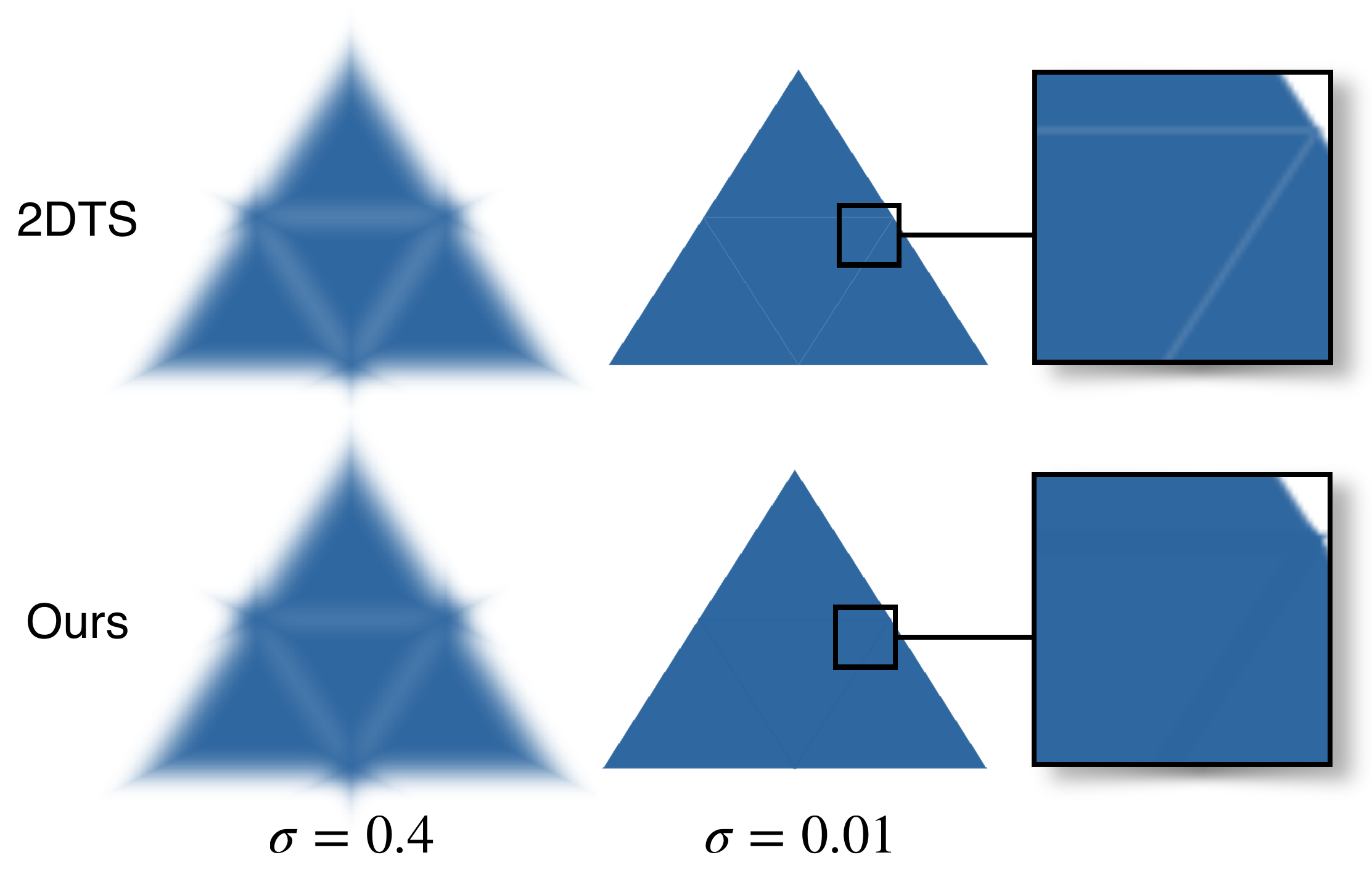}
  \caption{\textbf{Edge artifacts for connected triangles.} Transparent seams are unavoidable for soft kernels with high $\sigma$. However, only our elastic kernel produces seamless rendering with low $\sigma$ (hard boundary) thanks to its flexible value on the boundary. }
  \label{fig:edge_artifact}
\end{figure}

\paragraph{Bilateral Kernel.}
2DTS~\cite{sheng20252d} proposes a bilateral kernel function that extends beyond the triangle boundary:
\begin{equation}
G(x) = \exp\!\left(-\tfrac{1}{2}\,x^{\frac{2}{\sigma}}\right).
\end{equation}
By assigning non-zero weight for all $x$, this kernel maintains gradient flow in both interior and exterior regions. However, it has a fixed boundary value $G(1) = \exp(-\tfrac{1}{2}) \approx 0.61$. Even as $\sigma \to 0$ to sharpen the kernel, the boundary value remains 0.61, creating edge artifacts (\cref{fig:edge_artifact}) since the accumulated opacity of adjacent faces is $0.61 + (1-0.61) \times 0.61 = 0.8479$. This fixed boundary weight also causes softened triangles to visually expand, biasing geometry when $\sigma$ is large.

\paragraph{Elastic Kernel.} In addition to bilateral support of the positional gradient, we find that flexible boundary weight is also important, leading to the following kernel:
\begin{equation}
G(x) = \exp\!\left(-\sigma\, x^{\frac{2}{\sigma}}\right).
\end{equation}
This formulation is close to the 2DTS bilateral kernel but differs in the boundary behavior. When $\sigma$ is large, $G(1)$ is moderate and gradient support is wide, robustly driving large-scale geometric movement in early training. As $\sigma$ decreases, $G(1)$ approaches one and the edge artifacts in \cref{fig:edge_artifact} diminish. This \emph{elastic} behavior---diffuse and wide-reaching initially, sharp and precise finally---simultaneously resolves the gradient insufficiency of single-sided kernels and the bias caused by a fixed boundary weight.

\subsection{Shape Reconstruction with Triangle Meshes}
\label{sec:shape_recon}
For shape reconstruction, we represent the target surface as a closed triangle mesh and deform it to match a set of posed RGB images. The mesh is initialized from a unit sphere and optimized end-to-end: the vertex position $\{\bm{v}_i\}$, per-face color $\{\bm{c}_i\}$, and per-face opacity $\{o_i\}$ are the primary learnable parameters, together with a single shared softness coefficient $\sigma$ applied uniformly across all faces. At each training step, the mesh is rendered by splatting the triangles via \cref{eq:render} while ignoring their connectivity. An $\mathcal{L}_1$ photometric loss propagates gradients from pixel colors back through the renderer to vertex positions, driving the mesh to conform to the target geometry. However, photometric supervision alone is insufficient: unconstrained optimization is prone to geometric degeneracies such as face self-intersections, fold-overs, and degenerate collapsed triangles. We therefore augment the photometric loss with geometric regularization terms that suppress these instabilities and preserve mesh validity throughout optimization.


\paragraph{Softness Regularization.}
During optimization, the softness coefficient $\sigma$ can easily collapse to zero when the photometric loss is minimized. Then, the elastic kernel degenerates to a step function and gradient support shrinks to an infinitesimal region around the triangle boundary, leading to vanished positional gradients. Therefore, we add a penalty on small $\sigma$:
\begin{equation}
\mathcal{L}_\sigma = \sigma^{-2},
\end{equation}
which preserves a meaningful differentiable region throughout optimization.

\paragraph{Edge Regularization.}
To prevent extreme surface distortion, we penalize long edges:
\begin{equation}
\mathcal{L}_e = \frac{1}{|E|} \sum_{e \in E} \| e \|_2,
\end{equation}
where $E$ denotes the mesh edges and $\| e \|_2$ their length.

\paragraph{Normal Consistency.}
We apply normal consistency regularization to penalize angular deviations between adjacent face normals, discourage fold-overs, and encourage smooth surfaces:
\begin{equation}
\mathcal{L}_n = \frac{1}{|F|} \sum_{\langle f_0, f_1 \rangle}
\left(1 - \bm{n}_0 \cdot \bm{n}_1\right),
\end{equation}
where $F$ is the set of adjacent face pairs and $\bm{n}_0$, $\bm{n}_1$ are the unit normals of neighboring triangles $\langle f_0, f_1 \rangle$.

\paragraph{Laplacian Smoothing.}
We apply Laplacian smoothing~\cite{nealen2006laplacian} to penalize each vertex's deviation from the centroid of its neighbors, promoting uniform vertex spacing and suppressing local surface noise:
\begin{equation}
\mathcal{L}_\text{lap} = \frac{1}{|V|} \sum_{i \in V} \left\| \bm{v}_i - \frac{1}{|\mathcal{N}(i)|} \sum_{j \in \mathcal{N}(i)} \bm{v}_j \right\|_2,
\end{equation}
where $V$ is the set of mesh vertices and $\mathcal{N}(i)$ denotes the neighbors of $\bm{v}_i$.

\paragraph{Training Objective.}
The objective combines the ${L_1}$ photometric loss $\mathcal{L}_c$ with geometric regularizers:
\begin{equation}
\mathcal{L}_\text{mesh} = \mathcal{L}_c + \gamma 
\left(\lambda_e \mathcal{L}_e
+ \lambda_n \mathcal{L}_n
+ \lambda_\text{lap} \mathcal{L}_\text{lap}
+ \lambda_\sigma \mathcal{L}_\sigma\right),
\end{equation}
where $\gamma$ follows a cosine decay schedule, initialized large to stabilize early deformation and reduced gradually to allow fine geometric detail to emerge.


\subsection{Novel-View Synthesis with Triangle Soup}
\label{sec:nvs}
For novel-view synthesis, we represent the scene as an unstructured triangle soup. Each triangle is an independent primitive with its own learnable vertex positions, opacity, color (represented as spherical harmonic coefficients), and softness coefficient $\sigma$. The primitive count is not fixed: triangles are dynamically added and removed throughout training to adapt scene coverage. Supervision comes from posed multi-view RGB images, and the model is trained end-to-end with a photometric loss and regularization.

\paragraph{Loss Functions.}
The photometric loss $\mathcal{L}_c$ combines an ${L}_1$ term with a D-SSIM term~\cite{kerbl20233d}. Following Triangle Splatting~\cite{held2025triangle}, we include four auxiliary losses: an opacity loss $\mathcal{L}_o$~\cite{kheradmand20243d} that encourages low-opacity primitives, a distortion loss $\mathcal{L}_d$ that concentrates alpha weights along each ray for sharper renderings, a normal consistency loss $\mathcal{L}_n$~\cite{Huang2DGS2024} that aligns each triangle's normal with the surface normal estimated from depth map gradients, and a size regularization $\mathcal{L}_s$~\cite{held2025triangle} that encourages larger triangles to reduce over-fragmentation:
\begin{equation}
\mathcal{L}_\text{nvs} =
\mathcal{L}_c
+ \lambda_o \mathcal{L}_o
+ \lambda_d \mathcal{L}_d
+ \lambda_n \mathcal{L}_n
+ \lambda_s \mathcal{L}_s.
\end{equation}


\section{Experiments}
\label{sec:exp}

We evaluate our method on two tasks: shape reconstruction and novel-view synthesis (NVS) from multi-view images. In both settings, we compare kernel function variants within a common triangle splatting framework to isolate the effect of kernel design. Shape reconstruction is the more diagnostic setting for kernel behavior because the optimization starts from a fixed-topology sphere, and cannot rely on SfM initialization, densification, or pruning to compensate for poor gradient flow. NVS then tests whether the same kernel benefits persist in a fully adaptive splatting pipeline.


We evaluate shape reconstruction on the FAMOUS dataset~\cite{odedstein-meshes} (data processing details in \cref{sec:famous_dataset}) and novel-view synthesis on Mip-NeRF~360~\cite{barron2022mipnerf360}, Tanks\&Temples~\cite{Knapitsch2017}, and Deep Blending~\cite{hedman2018deep}. 
We measure geometric accuracy by Chamfer Distance (CD) and report rendering quality by PSNR, SSIM, and LPIPS.


\begin{table}[t]
\caption{\textbf{Quantitative results of shape reconstruction} on the FAMOUS dataset~\cite{odedstein-meshes}. Our elastic kernel consistently outperforms splatting-based baselines under both $\lambda_\sigma$ settings, approaching the accuracy of Nvdiffrast-V, which additionally relies on triangle connectivity to interpolate vertex colors.}
\label{tab:famous}
\footnotesize

\renewcommand{\arraystretch}{0.9}  
\centering\setlength{\tabcolsep}{3pt}
\begin{tabular}{@{}lccccc@{}}
\toprule
& $\lambda_{\sigma}$ & PSNR$\uparrow$ & SSIM$\uparrow$ & LPIPS$\downarrow$ & CD($\times 10^{-3}$)$\downarrow$ \\ 
\midrule
Nvdiffrast-F~\cite{laine2020modular} & - & 35.99 & 0.983 & 0.031 & 1.775                                               \\
Nvdiffrast-V~\cite{laine2020modular} & -  & \textbf{40.95} & \textbf{0.995} & \textbf{0.008} & \textbf{0.469}                                       \\
 \midrule
Triangle Splatting~\cite{held2025triangle} & $10^{-3}$ & 37.15 & 0.983 & 0.040 & 2.989                                                \\
2DTS~\cite{sheng20252d} & $10^{-3}$               & \underline{38.43} & \underline{0.987} & \underline{0.038} & \underline{0.926}                                                \\
Ours & $10^{-3}$  & \textbf{40.67} & \textbf{0.991} & \textbf{0.023} & \textbf{0.717}                                        \\ \midrule
Triangle Splatting~\cite{held2025triangle} & $10^{-4}$ & 29.06 & 0.958 & 0.068 & 5.369                                                \\
2DTS~\cite{sheng20252d} & $10^{-4}$               & \underline{40.13} & \underline{0.989} & \underline{0.034} & \underline{0.617}                                           \\
Ours & $10^{-4}$               & \textbf{40.90} & \textbf{0.991} & \textbf{0.023} & \textbf{0.598}                                       \\ \bottomrule
\end{tabular}
\end{table}

\begin{figure}[t]
  \centering
  \includegraphics[width=1\linewidth]{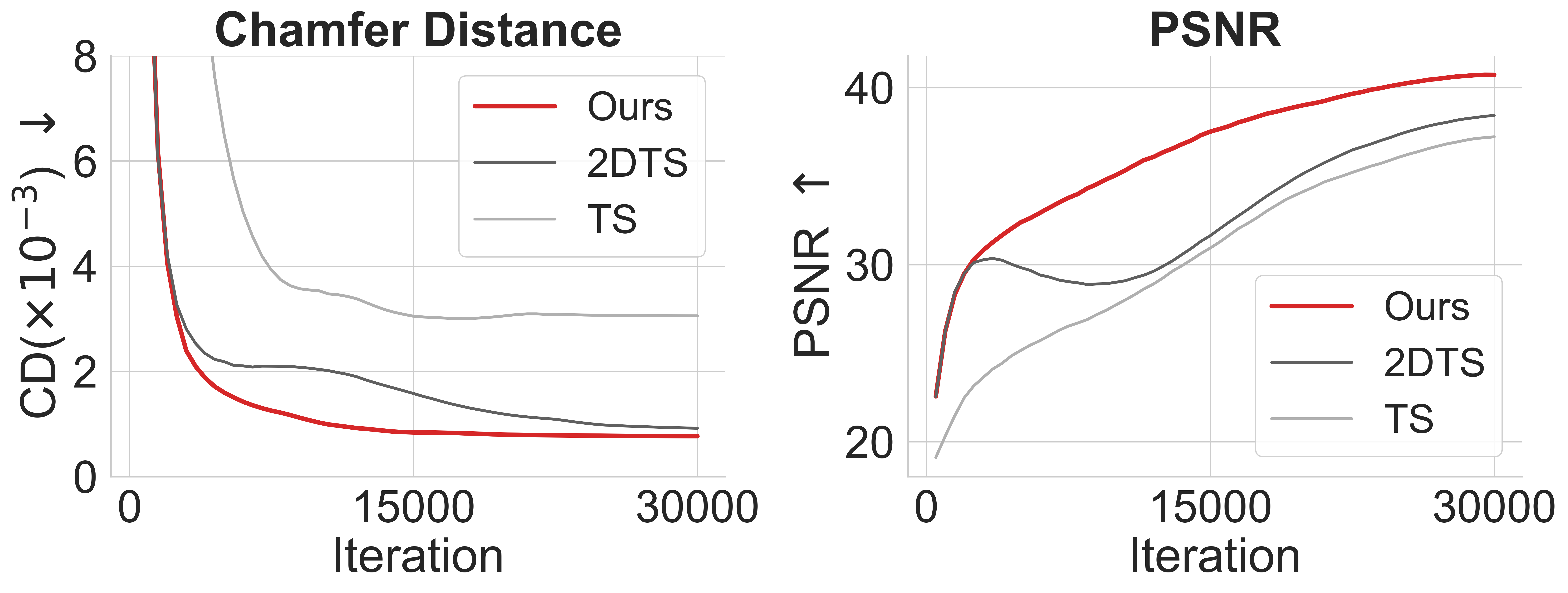}
  \caption{\textbf{Convergence speed comparison for mesh optimization.} We compare our elastic kernel against the single-sided (TS) and bilateral (2DTS) kernels on shapes from the FAMOUS dataset, averaged over all objects reported in the paper.}
  \label{fig:metric_curve}
\end{figure}

\begin{figure*}[t]
  \centering
  \includegraphics[width=1\textwidth]{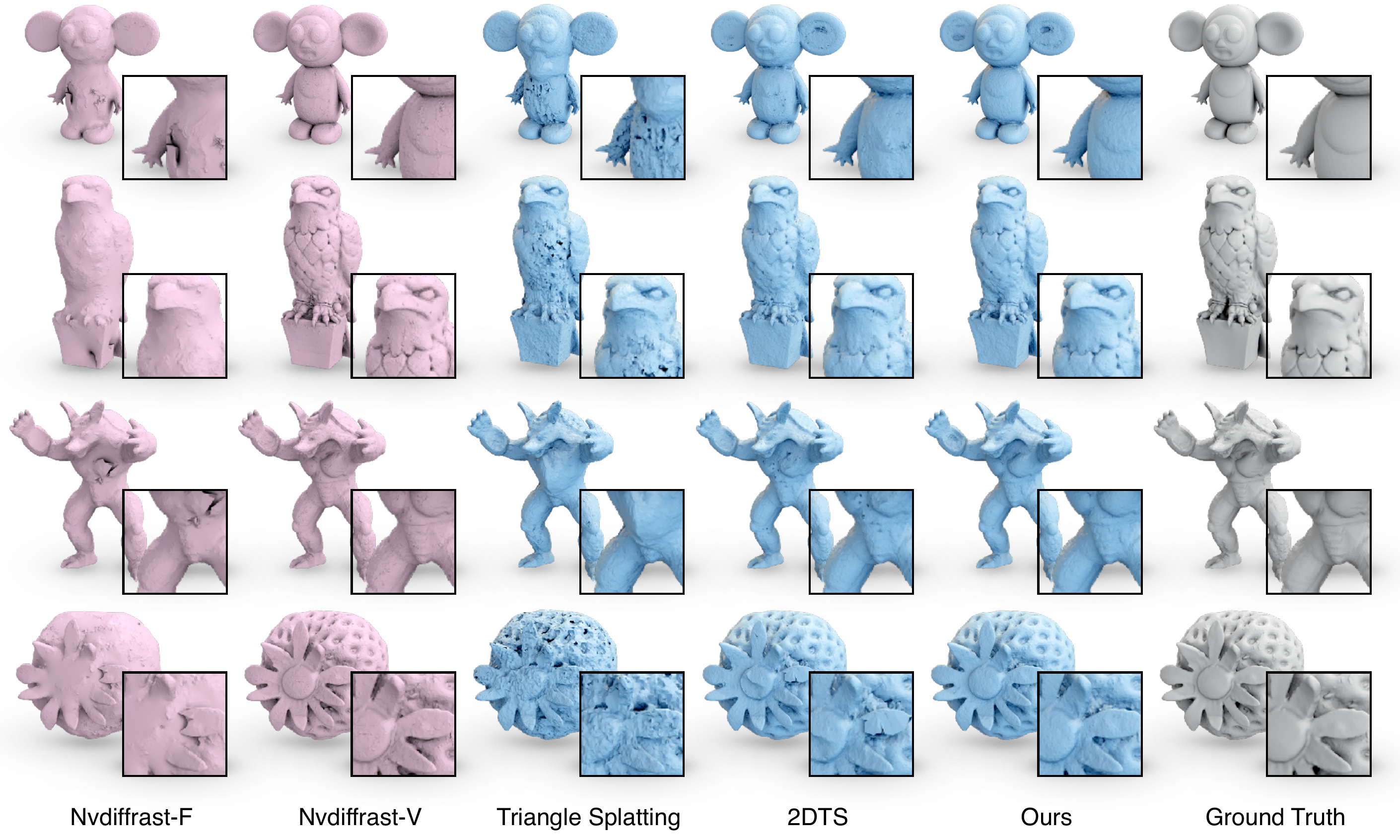}
  \caption{\textbf{Qualitative mesh reconstruction on the FAMOUS dataset~\cite{odedstein-meshes}.}
    Our elastic kernel recovers sharper surfaces and finer geometric details than Triangle Splatting~\cite{held2025triangle} and 2DTS~\cite{sheng20252d}.}
    \label{fig:mesh_famous}
\end{figure*}

\subsection{Shape Reconstruction}


\paragraph{Baselines.}
Within a unified framework, we compare our elastic kernel against the single-sided kernel of Triangle Splatting~\cite{held2025triangle} and the bilateral kernel of 2DTS~\cite{sheng20252d}.
We also include Nvdiffrast~\cite{laine2020modular} in two modes.
\emph{Nvdiffrast-F} (face-color mode) assigns a single constant color per face, making it the most direct analogue to splatting's per-primitive color; triangle interiors are non-differentiable with respect to vertex positions, so geometric gradients arise only from boundary antialiasing.
\emph{Nvdiffrast-V} (vertex-color mode) interpolates colors barycentrically across each face, providing strong interior gradients that effectively mitigate geometric error; we include it as an upper-bound reference rather than a direct competitor.


\begin{figure*}[t]
  \centering

\includegraphics[width=0.96\textwidth]{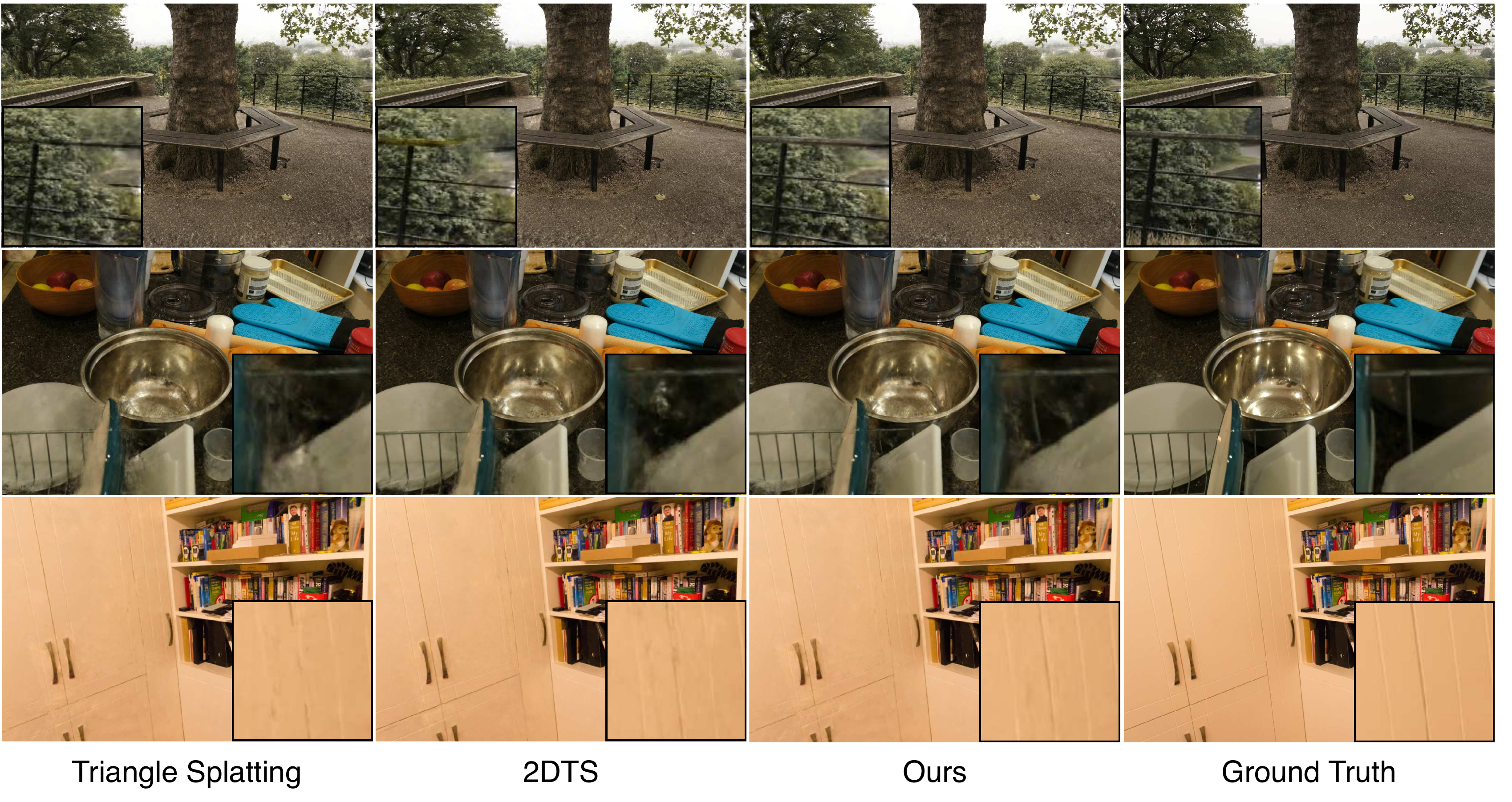}
  \caption{\textbf{Qualitative results of novel-view synthesis} on the Mip-NeRF 360~\cite{barron2022mipnerf360} and Tanks\&Temples~\cite{Knapitsch2017} datasets. Our kernel function results in better visual details and fewer floaters and artifacts.}
  \label{fig:nvs}
\end{figure*}

\paragraph{Results.} 
As shown in \cref{tab:famous} and \cref{fig:mesh_famous}, our method outperforms both splatting-based baselines on average across all objects and under both $\lambda_\sigma$ settings.
The gains are pronounced in all metrics, confirming that our kernel improves both geometric accuracy and rendering quality. Since all splatting variants use the same mesh initialization, training schedule, and regularization in this setting, the gap directly reflects improved optimization dynamics from the kernel design. Triangle Splatting shows larger errors in geometric details and tends to converge to less accurate solutions. 2DTS improves upon Triangle Splatting by recovering finer geometric details, although its reconstructed meshes still contain visible folds and cracks. In contrast, our elastic kernel produces cleaner surfaces and more accurate geometric details, consistent with the benefits of its bilateral gradient support and adaptive boundary behavior. As shown in \cref{fig:metric_curve}, the elastic kernel consistently converges faster and achieves a lower Chamfer Distance than Triangle Splatting and 2DTS throughout mesh optimization. Among the differentiable rasterization baselines, Nvdiffrast-F achieves a reasonable PSNR but substantially worse Chamfer Distance, revealing a clear gap between photometric and geometric reconstruction quality. Nvdiffrast-V, on the other hand, achieves the best overall numbers, but its per-vertex color interpolation provides additional appearance flexibility, making its results less directly comparable to the splatting-based methods.

\subsection{Novel-View Synthesis}

\paragraph{Baselines.}
We compare our kernel against the Triangle Splatting~\cite{held2025triangle} and 2DTS~\cite{sheng20252d} kernels within a unified framework. We additionally include 2DGS~\cite{Huang2DGS2024} and 3DGS~\cite{kerbl20233d} as Gaussian-based references.

\paragraph{Results.}
As shown in \cref{tab:nvs}, our method leads all triangle-based baselines in SSIM and LPIPS across every benchmark. The PSNR gap on Tanks\&Temples is marginal ($<$0.1~dB relative to TS), while perceptual metrics favor our approach throughout. As qualitatively demonstrated in \cref{fig:nvs}, our kernel function results in better visual details and fewer artifacts. Performance differences between kernel designs are smaller in NVS than in shape reconstruction. We attribute this to densification and pruning: adaptive topology control can partially offset suboptimal gradient flow, masking the intrinsic quality of individual kernels. Nevertheless, our elastic kernel consistently improves perceptual quality, confirming that a mathematically well-behaved kernel provides compounding benefits even alongside adaptive topology optimization. Against 3DGS, our method matches or exceeds SSIM and LPIPS on all three benchmarks despite using thin surface primitives rather than volumetric Gaussians. We attribute this to the adaptive $\sigma$ schedule: as optimization progresses, $\sigma$ decreases to sharpen triangle boundaries, enabling fine structural details to be captured without sacrificing early-stage convergence. The purpose of triangle splatting is not only to replace Gaussian primitives in image metrics, but also to optimize primitives that can form explicit surfaces. Unlike Gaussians, triangles can align and connect into a mesh, which is important for downstream editing, deformation, and physical simulation.

\begin{table}[t]
\setlength{\tabcolsep}{0.7pt}
\renewcommand{\arraystretch}{1.1}  
\scriptsize
\centering
\caption{\textbf{Quantitative results of novel-view synthesis} on Mip-NeRF~360~\cite{barron2022mipnerf360}, Tanks\&Temples~\cite{Knapitsch2017}, and Deep Blending~\cite{hedman2018deep}.
Our method is the most competitive among all triangle-based methods.}
\label{tab:nvs}
\begin{tabular}{@{}llllllllll@{}}
\toprule
     & \multicolumn{3}{c}{Mip-NeRF 360}                    & \multicolumn{3}{c}{Tanks\&Temples}                    & \multicolumn{3}{c}{Deep Blending}                   \\ 
 \cmidrule(l){2-4} \cmidrule(l){5-7} \cmidrule(l){8-10}
     & PSNR$\uparrow$ & SSIM$\uparrow$ & LPIPS$\downarrow$ & PSNR$\uparrow$ & SSIM$\uparrow$ & LPIPS$\downarrow$ & PSNR$\uparrow$ & SSIM$\uparrow$ & LPIPS$\downarrow$ \\ \midrule
3DGS~\cite{kerbl20233d} & \textbf{27.58} & \textbf{0.813} & 0.220             & \textbf{23.83} & 0.853          & 0.169             & \textbf{29.79} & \textbf{0.910}  & 0.238             \\
2DGS~\cite{Huang2DGS2024} & 26.82          & 0.794          & 0.260             & \underline{23.19}    & 0.833          & 0.212             & \underline{29.53}    & \underline{0.900}     & 0.256             \\
TS~\cite{held2025triangle}   & 27.02          & 0.804          & 0.200             & 23.10          & \underline{0.855}    & 0.144             & 29.02          & 0.891          & 0.242             \\
2DTS~\cite{sheng20252d} & 27.16          & \underline{0.808}    & \underline{0.194}       & 23.00          & 0.851          & \underline{0.143}       & 29.20          & 0.895          & \underline{0.229}       \\
Ours & \underline{27.33}    & \textbf{0.813} & \textbf{0.189}    & 23.08          & \textbf{0.858} & \textbf{0.134}    & 29.27          & 0.898          & \textbf{0.224}    \\ \bottomrule
\end{tabular}
\end{table}

\begin{figure}[!t]
  \centering
  \includegraphics[width=1\linewidth]{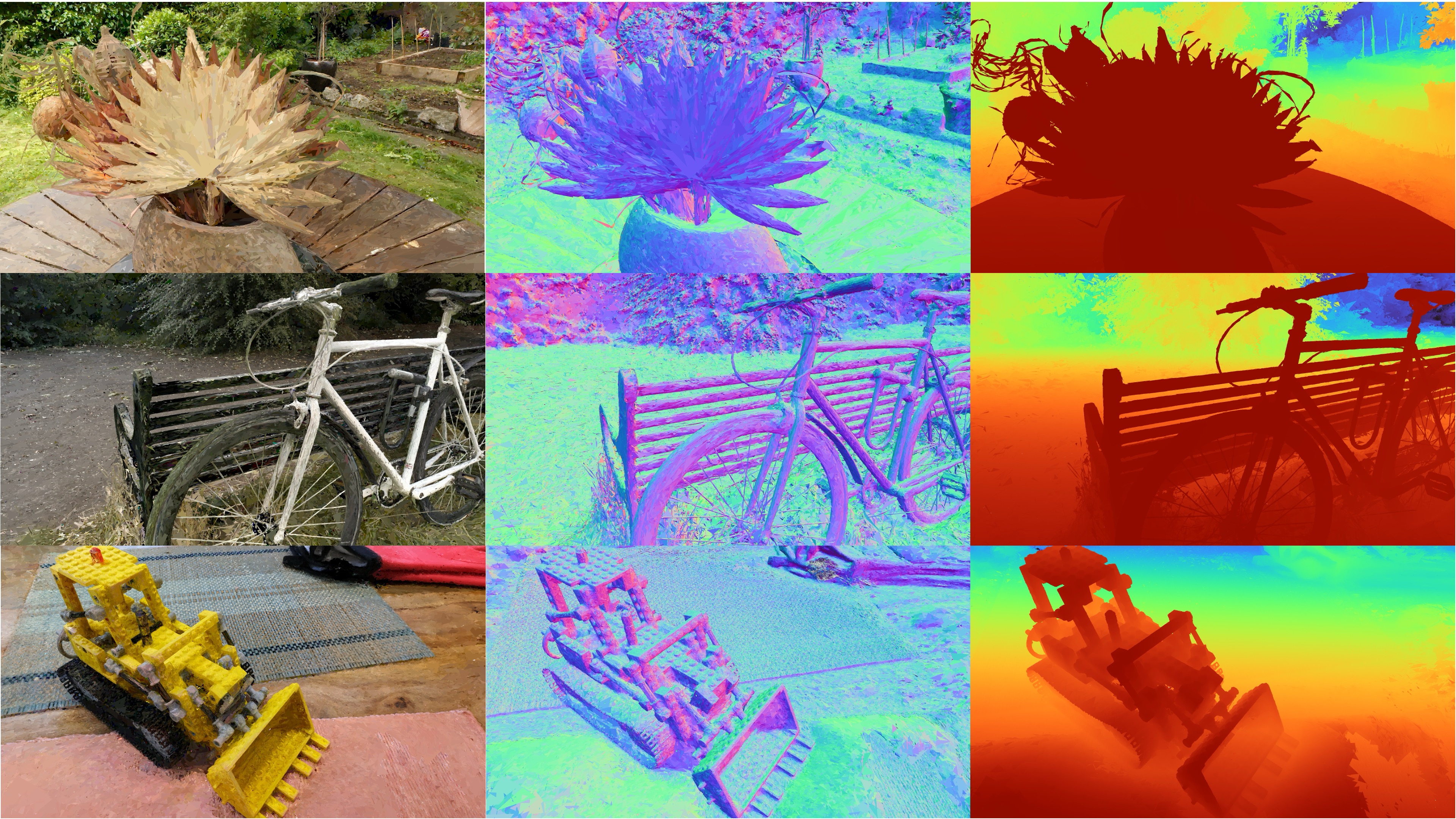}
    \caption{\textbf{Sharp and opaque rendering of optimized triangle soups} on the Garden, Bicycle, and Kitchen scenes from Mip-NeRF~360~\cite{barron2022mipnerf360}.
    }
  \label{fig:trisoup}
\end{figure}

Aiming to bridge splatting methods with the traditional graphics pipeline, we render the optimized triangle soup as sharp and opaque triangles, as shown by the examples in \cref{fig:trisoup}. As with Gaussian splats, very fine geometric detail still requires sufficient primitive density; the advantage of the elastic kernel is that sharper boundaries can be obtained without increasing the primitive count solely to compensate for boundary blur.

\section{Conclusion}
\label{sec:conclusion}

We presented Elastic Triangle Splatting, a unified differentiable rendering framework for connected meshes and unstructured triangle soups. Our elastic kernel combines wide bilateral gradient support with adaptive boundary opacity, enabling robust geometric optimization while producing clean boundaries as $\sigma$ decreases. Experiments on shape reconstruction and novel-view synthesis show consistent improvements over existing kernels, achieving state-of-the-art performance in both tasks.
\paragraph{Limitations.}
Our shape reconstruction pipeline assumes fixed mesh topology, limiting it to object-centric settings.
Very fine geometric details may also require many triangle primitives, similar to the primitive-density trade-off in Gaussian splatting.
Handling complex or open scenes would require topology-variant optimization~\cite{Held2025MeshSplatting} or hybrid representations~\cite{munkberg2022nvdiffrec,guedon2025milo}, for example combining a connected surface mesh with additional triangle splats for non-surface or high-frequency details.

{
    \small
    \bibliographystyle{ieeenat_fullname}
    \bibliography{main}
}

\clearpage

\section{Technical Details}

\subsection{Kernel Function Unification for Baselines}
\label{sec:kernel_unification}
In \cref{sec:kernel_func}, we reformulate all kernel functions using barycentric distance $x$ as a unified input. Here, we summarize their original forms.

In Triangle Splatting \cite{held2025triangle}, the kernel $G(s)=\operatorname{ReLU}(s)^{\sigma}$ is defined on a normalized variable $s\le 1$, where $s$ is the ratio between the pixel SDF (Signed Distance Function) and the SDF at the triangle incenter. Under this parameterization, $s>0$ indicates interior points, $s=0$ the boundary, and $s<0$ the exterior. We map this variable as $s=1-x$, which preserves mathematical consistency.

For 2DTS \cite{sheng20252d}, the original kernel is $G(x)=\exp(-0.5x^{2\sigma})$, where larger $\sigma$ produces a sharper profile, opposite to Triangle Splatting. We therefore re-parameterize the 2DTS kernel as $G(x)=\exp(-0.5x^{\frac{2}{\sigma}})$, so smaller $\sigma$ yields sharper kernels.

For shape reconstruction, we use unified kernel formulations to isolate the effect of kernel shape from the effect of distance parameterization. For Novel-View Synthesis (NVS), we keep the original baseline implementations to reflect state-of-the-art performance without unintended side effects from changing the distance metric.

\subsection{Adaptive Refinement}
\paragraph{Mesh Subdivision for shape reconstruction.}
To maximize the fidelity of the reconstruction mesh, we adaptively subdivide a group of triangles with the largest areas. For each selected triangle, we locate its longest edge and co-subdivide it with the neighboring triangle sharing that edge, yielding four smaller triangles while preserving topological consistency. Each triangle is processed at most once per subdivision step to avoid redundant refinements.
Subdivision is performed only at discrete refinement steps rather than at every optimization iteration, and it does not alter the differentiable rendering formulation. Between two subdivision steps, the mesh connectivity is fixed and only geometry, appearance, opacity, and softness are optimized. We use subdivision to increase final reconstruction fidelity; it is not required for the kernel comparison itself, and the final output remains a triangle mesh compatible with standard rasterization.

\paragraph{Densification and Pruning for NVS.}
We follow the densification strategy of 3DGS MCMC~\cite{kheradmand20243d} as adopted in Triangle Splatting. Triangles with larger opacity or smaller $\sigma$ (i.e., sharper, more surface-like primitives) are sampled with higher probability for cloning or subdivision. Triangles are dynamically cloned, split, and pruned throughout training based on their contribution to image reconstruction, allowing the representation to concentrate capacity in high-error or under-covered regions.

\subsection{FAMOUS Dataset Processing}
\label{sec:famous_dataset}
We evaluate shape reconstruction using ground-truth meshes from the FAMOUS dataset~\cite{odedstein-meshes}, spanning 21 diverse objects.
Multi-view training data is rendered at $1024\!\times\!1024$ pixels with 32 training views and 8 test views per object.
Cameras are placed on a spherical grid of 5 elevation angles ($-70^\circ$ to $70^\circ$) and 8 uniformly spaced azimuths, ensuring full object coverage.
Each scene is lit by two opposing directional lights along the Z-axis, with ambient and diffuse coefficients of 0.4 and 0.6 respectively; specular reflection is disabled to focus evaluation on geometric shading.

\subsection{Implementation Details}
\paragraph{Shape Reconstruction.}
All methods are initialized from an icosphere (subdivision level 4) whose scale and center are normalized to encompass the target object.
Shape reconstruction runs for 30{,}000 iterations. The subdivision process is deactivated once the mesh reaches 100{,}000 faces.
Regularization coefficients for edge, normal, and Laplacian smoothness are 1.5, 0.01, and 0.5, respectively.
The sigma regularization weight $\lambda_\sigma \in \{10^{-3}, 10^{-4}\}$ is applied to all splatting methods to compare kernel robustness under different levels of softness pressure.
The global regularization weight $\gamma$ follows a cosine decay from 1.0 at step 500 to 0 at step 30{,}000.
Background colors are randomized during training to discourage overfitting to a fixed scene prior. 
All experiments use a single NVIDIA L40S GPU.

\paragraph{Novel-View Synthesis.}
Triangle primitives are initialized from the sparse point cloud produced by structure-from-motion (SfM). All hyperparameters follow the original Triangle Splatting~\cite{held2025triangle} implementation.
All triangle-based methods use the hyperparameters of the original Triangle Splatting codebase. 2DGS and 3DGS use their respective default settings. All experiments run on a single NVIDIA L40S GPU.

\section{Additional Experiments}

\begin{table*}[t]
\centering
\caption{\textbf{Ablation study on small triangle filtering}. Not culling small triangles has negligible effect on novel-view synthesis performance.}
\label{tab:filter}
\footnotesize
\begin{tabular}{@{}cccccccccc@{}}
\toprule
    \multirow{2}{*}{cull small tri.} & \multicolumn{3}{c}{Mip-NeRF 360}                    & \multicolumn{3}{c}{Tanks\&Temples}                  & \multicolumn{3}{c}{Deep Blending}                   \\
    \cmidrule(l){2-4} \cmidrule(l){5-7} \cmidrule(l){8-10}
     & PSNR$\uparrow$ & SSIM$\uparrow$ & LPIPS$\downarrow$ & PSNR$\uparrow$ & SSIM$\uparrow$ & LPIPS$\downarrow$ & PSNR$\uparrow$ & SSIM$\uparrow$ & LPIPS$\downarrow$ \\
    \midrule
\checkmark  & \textbf{27.33} & \textbf{0.813} & \textbf{0.189}    & 23.08          & 0.858          & 0.134             & 29.27          & 0.898          & \textbf{0.224}    \\
  & 27.32          & \textbf{0.813} & 0.190             & \textbf{23.12} & \textbf{0.859} & \textbf{0.133}    & \textbf{29.28} & \textbf{0.900}  & \textbf{0.224}    \\ \bottomrule
\end{tabular}
\end{table*}

\subsection{Triangle Size Filtering}
In the original Triangle Splatting framework, a triangle-size filter in the CUDA kernel culls triangles that are too large or too small (e.g., sub-pixel primitives).
We empirically find that filtering out small triangles can be harmful, especially in mesh optimization: as subdivision makes the mesh finer, this filtering can introduce rendering holes and prevent accurate surface alignment. We therefore disable culling of small triangles. For NVS performance, we show in \cref{tab:filter} that keeping small triangles has negligible impact.

\subsection{Regularization Sensitivity}
To complement the shape reconstruction results in the main paper, we vary one non-$\sigma$ regularization weight at a time on a FAMOUS subset. As shown in \cref{tab:reg_ablation}, the original setting performs best. Moderate changes to the normal and Laplacian terms preserve the same overall behavior, while large edge or Laplacian weights over-smooth the mesh and degrade Chamfer Distance.

\begin{table}[!h]
\caption{\textbf{Regularization sensitivity} on a FAMOUS subset. Values are Chamfer Distance ($\times 10^{-3}$)$\downarrow$ after changing one regularization weight at a time.}
\label{tab:reg_ablation}
\centering
\footnotesize
\setlength{\tabcolsep}{15pt}
\begin{tabular}{@{}lccc@{}}
\toprule
Scale & $\lambda_e$ & $\lambda_n$ & $\lambda_\text{lap}$ \\
\midrule
$0.1\times$ & 0.891 & 0.648 & 0.694 \\
$1\times$ & \textbf{0.607} & \textbf{0.607} & \textbf{0.607} \\
$10\times$ & 2.097 & 0.666 & 1.935 \\
\bottomrule
\end{tabular}
\end{table}

\subsection{Runtime Efficiency}
In \cref{tab:efficiency}, we report the average optimization time per scene, triangle count, FPS, and rendering time per 1 million triangles on the Mip-NeRF 360 dataset.
The elastic kernel introduces a modest training-time overhead, while keeping triangle count and rendering throughput comparable to prior triangle kernels.

\begin{table}[ht]
\caption{\textbf{Efficiency comparison} on Mip-NeRF~360. Rendering time is measured per 1 million triangles.}
\label{tab:efficiency}
\centering
\setlength{\tabcolsep}{3.7pt}
\footnotesize
\begin{tabular}{@{}lcccc@{}}
\toprule
 & training time$\downarrow$ & \# triangles & FPS$\uparrow$ & rendering time $\downarrow$  \\
 \midrule
Triangle Splatting  & 58.54 min & 3.71M & 61.9 & 4.3 ms \\
2DTS                & 64.23 min & 3.58M & 58.4 & 4.8 ms \\ 
Ours                & 68.19 min & 3.72M & 57.5 & 4.7 ms \\
\bottomrule
\end{tabular}
\end{table}

\begin{table}[ht]
\centering
\scriptsize
\setlength{\tabcolsep}{2pt}
\renewcommand{\arraystretch}{1.2}
\caption{CD($\times 10^{-3}$)$\downarrow$ on FAMOUS~\cite{odedstein-meshes}.}
\label{tab:famous_cd}
\resizebox{\linewidth}{!}{%
\begin{tabular}{lcccccccc}
\toprule
\multicolumn{1}{c}{}                   & Nvdiffrast-F          & Nvdiffrast-V          & TS & 2DTS      & Ours      & TS & 2DTS      & Ours       \\
\cmidrule(l){2-3}
\cmidrule(l){4-6}
\cmidrule(l){7-9}
\multicolumn{1}{c}{$\lambda_{\sigma}$} & \multicolumn{1}{c}{-} & \multicolumn{1}{c}{-} & $10^{-3}$          & $10^{-3}$ & $10^{-3}$ & $10^{-4}$          & $10^{-4}$ & $10^{-4}$ \\ \midrule
armadillo                              & 0.14                     & 0.061                    & 1.649                         & 0.084                         & 0.079                         & 7.789                         & 0.079                         & 0.077                         \\
boot                                   & 1.515                    & 0.082                    & 0.601                         & 0.631                         & 0.192                         & 2.373                         & 0.174                         & 0.129                         \\
brucewick                              & 3.463                    & 2.146                    & 11.177                        & 2.393                         & 2.613                         & 7.474                         & 2.381                         & 2.453                         \\
bunny\_hr                              & 0.782                    & 0.047                    & 2.362                         & 0.15                          & 0.066                         & 4.548                         & 0.072                         & 0.064                         \\
cat-low-resolution                     & 0.571                    & 0.05                     & 0.713                         & 0.436                         & 0.059                         & 1.123                         & 0.121                         & 0.056                         \\
cow-low-resolution                     & 0.29                     & 0.085                    & 0.916                         & 0.161                         & 0.102                         & 1.185                         & 0.104                         & 0.091                         \\
demosthenes                            & 1.982                    & 0.459                    & 1.509                         & 0.611                         & 0.559                         & 10.867                        & 0.273                         & 0.304                         \\
falconstatue                           & 1.194                    & 0.338                    & 1.495                         & 0.489                         & 0.878                         & 4.228                         & 0.402                         & 0.564                         \\
goathead                               & 0.591                    & 0.307                    & 2.565                         & 0.414                         & 0.331                         & 1.166                         & 0.333                         & 0.323                         \\
hand\_closed                           & 0.226                    & 0.094                    & 1.066                         & 0.21                          & 0.16                          & 4.661                         & 0.149                         & 0.141                         \\
house                                  & 12.41                    & 4.623                    & 7.605                         & 6.576                         & 5.638                         & 10.018                        & 5.468                         & 5.488                         \\
human\_neutral                         & 0.433                    & 0.069                    & 0.307                         & 0.325                         & 0.079                         & 0.148                         & 0.086                         & 0.075                         \\
koala                                  & 0.167                    & 0.051                    & 1.928                         & 0.142                         & 0.058                         & 5.58                          & 0.059                         & 0.057                         \\
mushroom                               & 1.803                    & 0.061                    & 0.545                         & 0.433                         & 0.072                         & 2.285                         & 0.071                         & 0.067                         \\
nefertiti-lowres                       & 0.102                    & 0.055                    & 1.429                         & 0.091                         & 0.101                         & 5.715                         & 0.079                         & 0.099                         \\
scorpion                               & 0.119                    & 0.048                    & 0.585                         & 0.132                         & 0.077                         & 6.433                         & 0.074                         & 0.076                         \\
skull                                  & 7.675                    & 0.145                    & 7.286                         & 2.972                         & 0.548                         & 14.062                        & 1.062                         & 0.762                         \\
springer                               & 2.242                    & 0.545                    & 3.133                         & 1.764                         & 2.519                         & 6.88                          & 1.074                         & 0.871                         \\
strawberry                             & 0.502                    & 0.051                    & 12.486                        & 0.118                         & 0.069                         & 0.492                         & 0.076                         & 0.069                         \\
stuffedtoy                             & 0.546                    & 0.232                    & 2.251                         & 0.355                         & 0.303                         & 11.103                        & 0.268                         & 0.25                          \\
tree\_closed                           & 0.53                     & 0.305                    & 1.165                         & 0.957                         & 0.549                         & 4.624  & 0.554                         & 0.545\\
\bottomrule
\end{tabular}%
}
\end{table}

\begin{table}[t!]
\centering
\scriptsize
\setlength{\tabcolsep}{2.6pt}
\caption{PSNR$\uparrow$ on Mip-NeRF 360~\cite{barron2022mipnerf360}.}
\label{tab:mipnerf_psnr}
\begin{tabular}{@{}lccccccccc@{}}
\toprule
     & bicycle        & bonsai         & counter        & flowers        & garden         & kitchen        & room           & stump          & treehill       \\
     \midrule
3DGS & \textbf{25.15} & 32.48          & \underline{29.18}    & \textbf{21.41} & \underline{27.37}    & \textbf{31.56} & \textbf{31.90} & \textbf{26.69} & \textbf{22.50} \\
2DGS & 24.61          & 31.38          & 28.15          & \underline{20.81}    & 26.66          & 30.28          & 30.83          & \underline{26.19}    & \underline{22.44}    \\
TS   & 24.54          & 32.13          & 28.90          & 20.47          & 27.00          & 31.42          & 31.10          & 26.00          & 21.61          \\
2DTS & 24.70          & \underline{32.63}    & 28.99          & 20.56          & 27.19          & \underline{31.50}    & 31.44          & 25.91          & 21.54          \\
Ours & \underline{24.92}    & \textbf{32.79} & \textbf{29.24} & 20.79          & \textbf{27.42} & 31.41          & \underline{31.77}    & 26.10          & 21.53      
      \\
\bottomrule
\end{tabular}
\end{table}

\begin{table}[t!]
\centering
\scriptsize
\setlength{\tabcolsep}{2.6pt}
\caption{SSIM$\uparrow$ on Mip-NeRF 360~\cite{barron2022mipnerf360}.}
\label{tab:mipnerf_ssim}
\begin{tabular}{@{}lccccccccc@{}}
\toprule
      & bicycle              & bonsai         & counter        & flowers        & garden         & kitchen              & room           & stump          & treehill       \\ \midrule
3DGS & \underline{0.748} & 0.948          & 0.916          & 0.588 & \underline{0.857}    & \underline{0.933} &0.928 & \textbf{0.768} & \textbf{0.635} \\
2DGS & 0.713                & 0.936          & 0.900          & 0.554          & 0.834          & 0.921                & 0.915          & 0.753          & \underline{0.619}    \\
TS   & 0.738                & 0.949          & 0.914          & 0.585          & 0.849          & 0.932                & 0.930          & 0.744          & 0.593          \\
2DTS & 0.743                & \underline{0.952}    & \underline{0.917}    & \underline{0.590}    & 0.852          & \underline{0.933}          & \underline{0.933}    & 0.751          & 0.603          \\
Ours & \textbf{0.754}       & \textbf{0.954} & \textbf{0.920} & \textbf{0.597} & \textbf{0.858} & \textbf{0.935}       & \textbf{0.936} & \underline{0.758}    & 0.608\\
\bottomrule
\end{tabular}
\end{table}

\begin{table}[t!]
\centering
\scriptsize
\setlength{\tabcolsep}{2.6pt}
\caption{LPIPS$\downarrow$ on Mip-NeRF 360~\cite{barron2022mipnerf360}.}
\label{tab:mipnerf_lpips}
\begin{tabular}{@{}lccccccccc@{}}
\toprule
      & bicycle              & bonsai         & counter        & flowers        & garden         & kitchen              & room           & stump          & treehill     \\ \midrule
3DGS & 0.241          & 0.180          & 0.182          & 0.359          & 0.122          &0.116 & 0.195 & 0.243 & 0.346 \\
2DGS & 0.306          & 0.204          & 0.214          & 0.404          & 0.163          & 0.138          & 0.222          & 0.290          & 0.398          \\
TS   & 0.224          & 0.143          & 0.157          & 0.316          & 0.122          & 0.107          & 0.163          & 0.242          & 0.322          \\
2DTS & \underline{0.220}    & \textbf{0.137} & \underline{0.150}    & \underline{0.307}    & \underline{0.117}    & \underline{0.106}    & \underline{0.161}    & \underline{0.235}    & \underline{0.317}    \\
Ours & \textbf{0.208} & \underline{0.138}    & \textbf{0.148} & \textbf{0.299} & \textbf{0.111} & \textbf{0.105} & \textbf{0.160} & \textbf{0.226} & \textbf{0.311}\\
\bottomrule
\end{tabular}
\end{table}

\subsection{Per-scene Results}
In \cref{tab:famous_cd}, we provide a detailed comparison of the Chamfer Distance (CD) across 21 diverse objects from the FAMOUS dataset~\cite{odedstein-meshes}.
\cref{tab:mipnerf_psnr,tab:mipnerf_ssim,tab:mipnerf_lpips,tab:ttdb} 
show the per-scene results on Mip-NeRF 360~\cite{barron2022mipnerf360}, Tanks\&Temples~\cite{Knapitsch2017} and Deep Blending~\cite{hedman2018deep} datasets.

\begin{table*}[!t]
\footnotesize
\centering
\caption{\textbf{Per-scene results of novel-view synthesis} on Tanks\&Temples~\cite{Knapitsch2017} and Deep Blending~\cite{hedman2018deep}.}
\centering
\label{tab:ttdb}
\begin{tabular}{@{}lcccccccccccc@{}}
\toprule
 & \multicolumn{3}{c}{Train}                           & \multicolumn{3}{c}{Truck}                           & \multicolumn{3}{c}{Drjohnson}                       & \multicolumn{3}{c}{Playroom}                        \\
\cmidrule(l){2-4} \cmidrule(l){5-7} \cmidrule(l){8-10} \cmidrule(l){11-13}
       & PSNR$\uparrow$ & SSIM$\uparrow$ & LPIPS$\downarrow$ & PSNR$\uparrow$ & SSIM$\uparrow$ & LPIPS$\downarrow$ & PSNR$\uparrow$ & SSIM$\uparrow$ & LPIPS$\downarrow$ & PSNR$\uparrow$ & SSIM$\uparrow$ & LPIPS$\downarrow$ \\
       \midrule
3DGS                        & \textbf{22.21}       & \underline{0.82}     & 0.195             & \textbf{25.44} & \textbf{0.89}  & 0.142             & \textbf{29.49}       & \textbf{0.91}  & \underline{0.236}             & \underline{30.08}    & \textbf{0.91}  & 0.240             \\
2DGS                        & 21.26                & 0.79           & 0.250             & \underline{25.12}    & 0.87           & 0.173             & 28.90                & \underline{0.90}     & 0.257             & \textbf{30.15} & \textbf{0.91}  & 0.260             \\
TS                          & 21.22                & \underline{0.82}     & 0.182             & 24.99          & \textbf{0.89}  & \textbf{0.105}    & 28.87                & 0.89           & 0.250             & 29.18          & 0.89           & 0.234             \\
2DTS                        & 21.24                & \underline{0.82}     & \underline{0.180}       & 24.77          & \underline{0.88}     & \underline{0.106}       & 28.81                & 0.89           & 0.240       & 29.58          & \underline{0.90}     & \underline{0.217}       \\
Ours                        & \underline{21.47} & \textbf{0.83}  & \textbf{0.162}    & 24.68          & \textbf{0.89}  & \textbf{0.105}    & \underline{28.91} & \underline{0.90}     & \textbf{0.235}    & 29.64          & \underline{0.90}     & \textbf{0.214}   
      \\
\bottomrule
\end{tabular}
\end{table*}

\end{document}